\pdfoutput=1

\documentclass[11pt]{article}

\usepackage[acceptedWithA]{tacl2021v1}
\usepackage{times}
\usepackage{latexsym}
\usepackage[T1]{fontenc}
\usepackage{url}
\usepackage{xspace,mfirstuc,tabulary}
\usepackage[utf8]{inputenc}
\usepackage{microtype}
\usepackage{hyperref}

\usepackage{amsmath,amsthm}
\usepackage{acronym}
\usepackage{booktabs} 
\usepackage{graphicx}
\usepackage[skip=2pt]{caption}
\usepackage{subcaption}
\usepackage{enumerate}
\usepackage[inline]{enumitem}

\usepackage{amsfonts}
\usepackage{amssymb}
\usepackage{mathbbol}
\usepackage{calligra}

\usepackage{algpseudocode} 
\usepackage{multirow}
\usepackage[linesnumbered,ruled]{algorithm2e}
\usepackage{verbatim}
\usepackage{arydshln}
\newif\iftaclinstructions
\taclinstructionsfalse 
\iftaclinstructions
\renewcommand{\confidential}{}
\renewcommand{\anonsubtext}{(No author info supplied here, for consistency with
TACL-submission anonymization requirements)}
\newcommand{\instr}
\fi

\iftaclpubformat 

\else

\fi

\newcommand{\tabincell}[2]{\begin{tabular}{@{}#1@{}}#2\end{tabular}}

\newcommand{\myUparrow}{\color{red}{\Uparrow}}
\newcommand{\myuparrow}{\color{red}{\uparrow}}

\SetCommentSty{mycommfont}
\acrodef{TDS}{task-oriented dialogue system}
\acrodef{ODS}{open-domain dialogue system}
\acrodef{GRU}{gated recurrent unit}
\acrodef{RNN}{recurrent neural network}

\acrodef{ST}{self-training}
\acrodef{ICAST}{intent-calibrated self-training}

\acrodef{IGen}{intent generation}
\acrodef{InfoG}{information gain}
\acrodef{IAInfoG}{intent-aware answer information gain}
\acrodef{AInfoG}{answer information gain}
\acrodef{ICAS}{intent-calibrated answer selection}
\acrodef{ICA}{intent-calibration}

\acrodef{APL}{answer pseudo label}
\acrodef{IAPL}{intent-aware answer pseudo label}
\acrodef{IPL}{intent pseudo label}

\acrodef{BCELoss}{binary cross entropy loss}

\acrodef{OQ}{original question}
\acrodef{PA}{potential answer}
\acrodef{IR}{information request}
\acrodef{FD}{futher detail}
\acrodef{FQ}{following up question}
\acrodef{CQ}{clarification question}
\acrodef{PF}{positive feedback}
\acrodef{NF}{negative feedback}
\acrodef{GG}{greetings/gratitude}
\acrodef{JK}{junk}
\acrodef{O}{others}

\acrodef{NSP}{next sentence prediction}
\acrodef{MC dropout}{Monte-Carlo dropout}
\acrodef{PLM}{pretrained language model}
\acrodef{MAP}{mean average precision}
\acrodef{IAC}{intent-aware calibrator}
\acrodef{PLA}{pseudo label annotator}
\acrodef{TSST}{teacher-student self-training}

\acrodef{ICG}{intent confidence gain}
\acrodef{ICGE}{intent confidence gain estimation}
\acrodef{ALC}{answer label calibration}
\acrodef{IG}{intent generation}

\begin{document}
\title{Intent-calibrated Self-training for Answer Selection\\ in Open-domain Dialogues}

\author{
    \\
  \textbf{Wentao Deng}$^1$, \textbf{Jiahuan Pei}$^2$, \textbf{Zhaochun Ren}$^1$, \textbf{Zhumin Chen}$^1$, \textbf{Pengjie Ren}$^{1}$\thanks{~~Corresponding author.}
  \\
  $^1$Shandong University, Qingdao, China
  \\
  \texttt{wentao.deng@mail.sdu.edu.cn}
  \\
  \texttt{\{zhaochun.ren,chenzhumin,renpengjie\}@sdu.edu.cn}
  \\
  \\
  $^2$Centrum Wiskunde \& Informatica, Amsterdam, Netherlands
  \\
  \texttt{Jiahuan.Pei@cwi.nl}
}

\maketitle
\begin{abstract}
Answer selection in open-domain dialogues aims to select an accurate answer from candidates.
Recent success of answer selection models hinges on training with large amounts of labeled data. 
However, collecting large-scale labeled data is labor-intensive and time-consuming.
In this paper, we introduce the predicted intent labels to calibrate answer labels in a self-training paradigm. 
Specifically, we propose the \acf{ICAST} to improve the quality of pseudo answer labels through the intent-calibrated answer selection paradigm, in which we employ pseudo intent labels to help improve pseudo answer labels. 
We carry out extensive experiments on two benchmark datasets with open-domain dialogues.
The experimental results show that \ac{ICAST} outperforms baselines consistently with 1\%, 5\% and 10\% labeled data.
Specifically, it improves 2.06\% and 1.00\% of F1 score on the two datasets, compared with the strongest baseline with only 5\% labeled data.
\end{abstract}

\section{Introduction}

\Acfp{ODS} interact with users by dialogues in open-ended domains~\cite{huang2020challenges}. 
The responses in \ac{ODS} can be divided into different types, such as answer, gratitude, greeting and junk~\cite{qu2018analyzing}.
In this paper, we focus on selecting answers, which aims to identify the correct answer from a pool of candidates given a dialogue context.
Typically, there are two main branches of approaches to produce answers, i.e., generation-based methods and selection-based methods~\cite{park2022bert}. 
The former generate a response token by token; and the latter select a response from a pool of candidates. 
Currently, pure generation methods such as ChatGPT still face challenges: 
(1) They may generate incorrect contents. 
(2) They cannot generate timely answers. Thus, it still needs selection-based methods to improve the correctness and timeliness of generation-based method.

\begin{figure}[t!]
    \centering
    \includegraphics[width=\columnwidth]{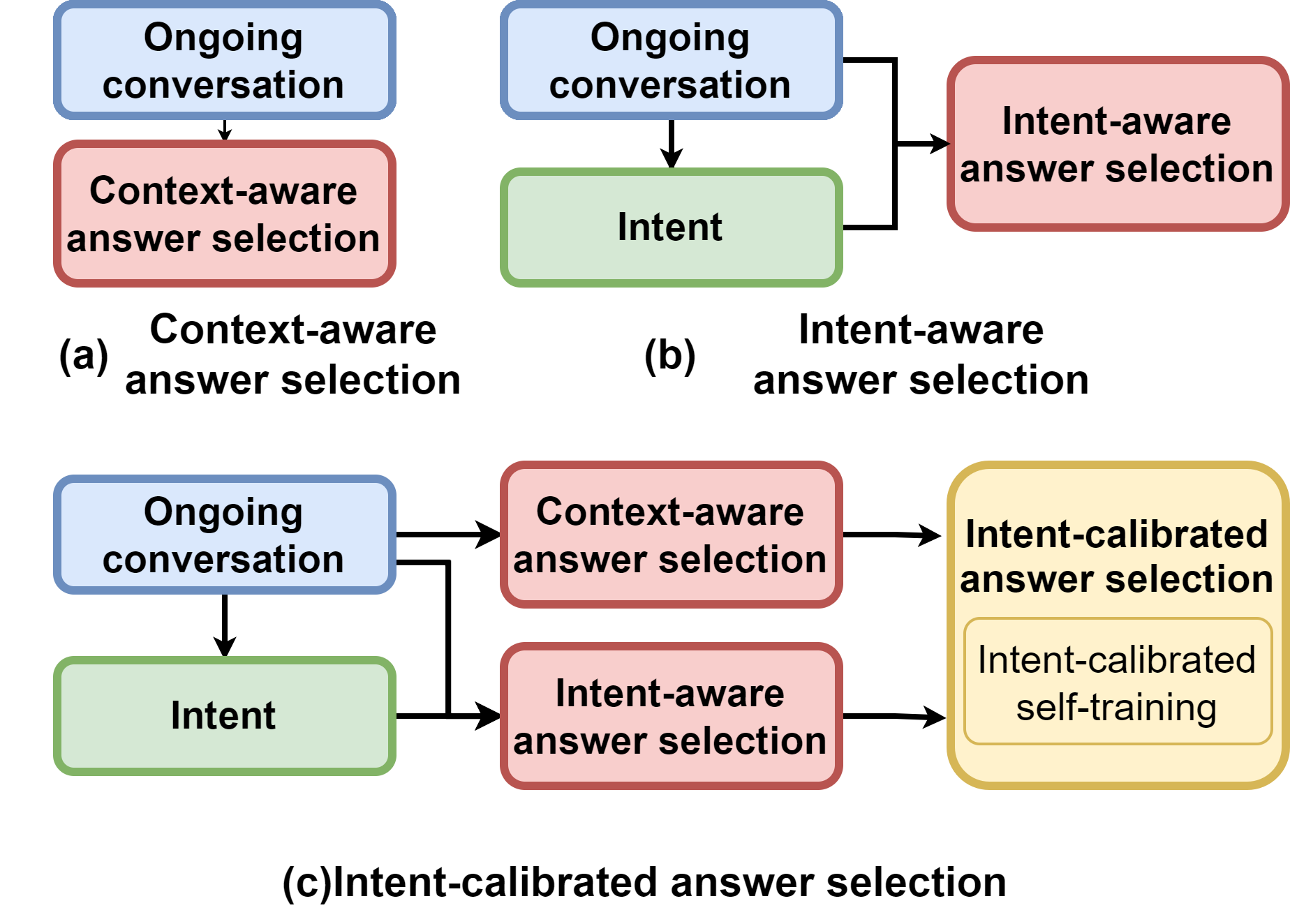}
    \caption{
    Comparison between previous answer selection models and our proposed framework.
    (a) Context-aware answer selection. 
    (b) Intent-aware answer selection. 
    (c) Intent-calibrated answer selection.
    }
    \label{fig:motivation}
\end{figure}

Figure \ref{fig:motivation} illustrates our idea by comparing the answer selection paradigms of (a) context-aware methods, (b) intent-aware methods, and (c) intent-calibrated methods. 
Context-aware methods (See Figure \ref{fig:motivation} (a)) capture the context of the ongoing dialogue for understanding users' information needs to select the most relevant responses from answer candidates~\cite{jeong2021label}. 
Unlike \aclp{TDS}, it is much more challenging for \acp{ODS} to infer users' information needs due to their open-ended goals~\cite{huang2020challenges}.

To this end, user intents, i.e., a taxonomy of utterances, are introduced to guide the information-seeking process~\cite{qu2018analyzing,qu2019user,yang2020iart}. 
If the intent of the previous \ac{OQ} is not satisfied by the \ac{PA} provided by a system, then the users' next intent is more likely to be \ac{IR}. 
For example, if the user asks: ``Can you send me a website, so I can read more information?'', the user's intent is \ac{IR}.
If the system does not consider the intent label \ac{IR}, then it may provide an answer which does not satisfy the user's request.

Intent-aware methods (See Figure \ref{fig:motivation} (b)) adopt intents as an extra input to better understand users' information needs in an utterance~\cite{yang2020iart}. 
However, they require sufficient human-annotated intent labels for training, the construction of which is time-consuming and labor-intensive. 

\Acl{ST} has been widely used to mitigate label scarcity problem~\cite{liu2022confidence, yang2022st++, zhang2022mdd}.
But it is still under-explored for answer selection in \acp{ODS}.
The principle of self-training is to iteratively learn a model by assigning pseudo-labels for large-scaled unlabeled data to extend the training set~\cite{amini2022self}.
The teacher-student self-training framework has been widely used in many recent works, where the teacher generates pseudo-labels and the student makes predictions~\cite{xie2020self,ghiasi2021multi,li2021metats,karamanolakis2021self}.
However, noisy pseudo labels incur error propagation across iterations, so the key challenge is how to assure both quality and quantity of pseudo labels~\cite{karamanolakis2021self}.

In this paper, we introduce an \acl{ICAS} paradigm, as in Figure~\ref{fig:motivation}(c).
It first conducts both context-aware and intent-aware answer selection to predict pseudo intent and answer labels, and then it selects high-quality intent labels to calibrate final answer labels.
To be more specific, we develop an \acf{ICAST} algorithm based on the teacher-student self-training and intent-calibrated answer selection paradigm.

The core procedure is:
First, we train a teacher model on the labeled data and predict pseudo intent labels for the unlabeled data. 
Second, we select high-quality intent labels by estimating intent confidence gain and then add selected intents to the input of the answer selection model. 
The intent confidence gain measures how much information a candidate intent label can bring to the model. 
Third, we re-train a student model on both the labeled and pseudo-labeled data.
Intuitively, \ac{ICAST} synthesizes pseudo intent and answer labels and integrates them into teacher-student self-training, which can assure synthetic answer quality by high-quality intents.

We conduct experiments on two datasets: \text{MSDIALOG}\footnote{\url{https://ciir.cs.umass.edu/downloads/msdialog/}} \cite{qu2018analyzing} and \text{MANTIS}\footnote{\url{https://github.com/guzpenha/mantis}}  \cite{penha2019introducing}.
The experimental results show that \ac{ICAST} outperforms the state-of-the-art baseline by 2.51\%/0.63\% of F1 score on \text{MSDIALOG}/\text{MANTIS} dataset, with 1\% labeled data.
The results demonstrate the effectiveness of \ac{ICAST} which selects accurate answers with incorporating high-quality predicted intent labels.

\section{Related Work}
In this section, we summarize related work in terms of three categories, i.e., traditional answer selection models, intent-aware answer selection models, and self-training for data argumentation.

\subsection{Traditional answer selection models} 
The dominant work focuses on modeling the representation of dialogue contexts, responses, and their relevance to select appropriate answers~\cite{zhou2016multi,zhou2018multi,chaudhuri2018improving}.
~\citet{wang2019multi} propose a sequential matching network to model the relation between the contextual utterances and the response by a cross-attention matrix.
~\citet{yang2019friendsqa} encode dialogue contexts and responses for answer utterance selection and answer span selection using multiple self-attention models, e.g., R-Net~\cite{wang2017gated} based on RNN and QANet~\cite{yu2018qanet} based on CNN.
Many researchers also explore to enhance the dialogue contexts or candidate responses.
~\citet{medvevd2020employing} extend the input candidate sentence with selected information from preceding sentence context.
~\citet{fu2020context} extend the contexts of the responses and integrate the context-to-context matching with context-to-response matching.
Several studies~\cite{ohmura2018context, barz2021incremental} also propose to improve the quality of answers by re-ranking answer candidates.

More recently, transformer-based pre-trained models have been the state-of-the-art paradigms~\cite{kim2019eighth,henderson2019polyresponse,tao2021survey}.
Researchers~\cite{henderson2019training,yang2019friendsqa} apply a BERT encoder~\cite{kenton2019bert} pre-trained on large-scaled open-domain dialogue corpus and fine-tune the model on small-scale in-domain dataset to capture the nuances. 
Likewise, ~\citet{whang2020effective} also use BERT encoder and perform context-response matching, but they also introduce the next utterance prediction and masked language modeling tasks during the post-training.
~\citet{gu2020speaker} incorporate speaker-aware embeddings into BERT to help with context understanding in multi-turn dialogues.
~\citet{liu2021filling} conduct utterance-aware and speaker-aware representations for dialogue contexts based on masking mechanisms in transformer-based pre-trained models, including BERT~\cite{kenton2019bert}, RoBERTa~\cite{liu2019roberta}, and ELECTRA~\cite{clark2019electra}.

There are several works which use auxiliary tasks to enhance answer selection. 
\citet{wu2020tod} incorporate a BERT-based response selection model with a contrastive learning objective and multiple auxiliary learning tasks, i.e., intention recognition, dialogue state tracking, and dialogue act prediction.
\citet{xu2021learning} enhance the response selection task with several auxiliary tasks, which can bring in extra supervised signals in multi-task learning manner.
\citet{pei2021cooperative} jointly learn missing user profiles with personalized response selection, which can improve response equality gradually based on enriched user profiles and neighboring dialogues.

\subsection{Intent-aware answer selection models}
Intent detection is a key prior to understand users' intent for answer selection, especially in multiple turn dialogues~\cite{gu2020speaker,park2022bert}.
Various deep NLP models have been adopted to classify intents~\cite{chen2017survey,liu2019review,weld2021survey,wang2021pre}.
~\citet{chen2016zero} generate new intents to bridge the semantic relation across domains for intent expansion and classification. 
~\citet{wu2020tod} improve pre-trained BERT with an extra contrastive objective for intention recognition.
The key challenge is natural language understanding with the state-of-the-art NLP models, e.g., CNNs~\cite{chen2016zero}, RNNs~\cite{firdaus2021deep}, transformers~\cite{zhao2020condition}, \acp{PLM}~\cite{wu2020tod,yan2022remedi}.

Intent calibration researches attempt to predict additional information to resolve users' ambiguous or uncertain intents.
\citet{lin2019post} calibrate the confidence of the softmax outputs for unknown intent detection. 
\citet{gong2022confidence} represent labels in hyperspherical space uniformly and calibrate confidence to trade-off accuracy and uncertainty. 
However, none of the above works has adapted the detected intents to answer selection. 
The most related work is IART~\cite{yang2020iart}, which weights the context by attending predicted intents for response selection. 

Unlike above methods, we propose to improve the performance of answer selection by using large amount of unlabeled data.
We devise the intent-calibrated self-training to improve the quality of pseudo answer labels by considering user intents.

\subsection{Self-training for answer selection}
Self-training has received remarkable attention in natural language processing~\cite{luo2022self} and machine learning~\cite{karamanolakis2021self,amini2022self}.
In general, the core idea is to augment the model training with pseudo supervision signals~\cite{wu2020tod,yan2022remedi}. 

~\citet{sachan2018self} introduce a self-training algorithm for jointly learning to answer and generate questions, which augments labeled question-answer pairs with unlabeled text.
\citet{wu2018learning} introduce a pre-trained sequence-to-sequence model as an annotator to generate pseudo labels for unlabeled data to supervise the training process.
\citet{deng2021learning} propose to use the fine-tuned question generator and answer generator to generate pseudo question-answer pairs.
\citet{lin2020world} introduce a fine-tuned generation-base model to generate gray-scale data.

Differently, the proposed \ac{ICAST} in this work seeks to improve the quality of pseudo answer labels by introducing the intent-calibrated pseudo labeling mechanism which uses high-quality pseudo intent label to calibrate pseudo answer labels.
\section{Preliminary}
\subsection{Answer selection task}

We form answer selection as a binary classfication task~\cite{yang2020iart}. 
We denote the labeled dataset $D^l=\{([\mathbf{x}_{i},\mathbf{e}_{i}], y_i)\}\vert_{i=1}^{\vert D^l \vert}$ and unlabeled dataset $D^u=\{\mathbf{x}_i\}\vert_{i=1}^{\vert D^u \vert}$.
For the $i$-th sample, $\mathbf{x}_{i}=(\mathbf{u}_i,a_i)$ is a context-candidate pair, which consists of as a sequence of utterances as the context $\mathbf{u}_{i}=[u_1, \cdots, u_{|\mathbf{u}_{i}|}]$ and a candidate answer $a_i \in A$ (a set of all candidate answers). 
$\mathbf{e}_i=[e_1,\cdots,e_{|\mathbf{u}_i|}]$ is a sequence of user intent labels.
$y_i \in \{0, 1\}$ is the answer label, $y_i=1$ denotes $a_i$ is a correct answer, otherwise $y_i=0$.

Our task is to learn a model $f=[f^{\alpha}, f^{\beta}]$.
The intent generation module $f^{\alpha}$ predicts a set of intents $\tilde{\mathbf{e}}_{i}$ given a context-candidate pair, parameterized by $\alpha$;
The answer selection module $f^{\beta}$ predicts an answer label given context and the predicted intents, parameterized by $\beta$.
Formally, we estimate the following probabilities:
\begin{align}
     f^{\alpha} &:= P(\mathbf{e}_{i}|\mathbf{u}_{i}, a_{i}, \alpha),\\
     f^{\beta} &:= P(y_i \vert \mathbf{u}_{i}, \tilde{\mathbf{e}}_{i}, a_{i}, \beta).
\end{align}

\subsection{BERT for answer selection}
BERT~\cite{kenton2019bert} is widely used to model the semantic dependency between context and candidate answers in recent researches~\cite{qu2019bert, li2019bertsel,matsubara2020reranking,yang2020iart}. 
First, we format the input of BERT as $\mathbf{x}_i=[[CLS];\mathbf{u}_i;[SEP];a_i]$, where the special token $[CLS]$ indicates the beginning of a context-candidate pair, and $[SEP]$ is a separator. 
Then, we use BERT to encode $\mathbf{x}_i$ and get the representation of $[CLS]$ token $\mathbf{h}_i^{CLS}$.
Next, let $\mathbf{h}_i^{CLS}$ pass through a linear layer followed by an activation function to compute the probability $p_i$ of a candidate answer. Formally,
\begin{align}    
    p_i &=\sigma(\mathbf{W}\mathbf{h}_i^{CLS}+\mathbf{b}), \label{p-cls} \\
    \mathbf{h}_i^{CLS} &= BERT(\mathbf{x}_i), \label{h-cls}
\end{align}
where $\mathbf{W}$ and $\mathbf{b}$ are trainable parameters, $\sigma$ is sigmoid function.

\begin{figure}[tp]
    \label{method-figure}
    \centering
    \includegraphics[width=8cm]{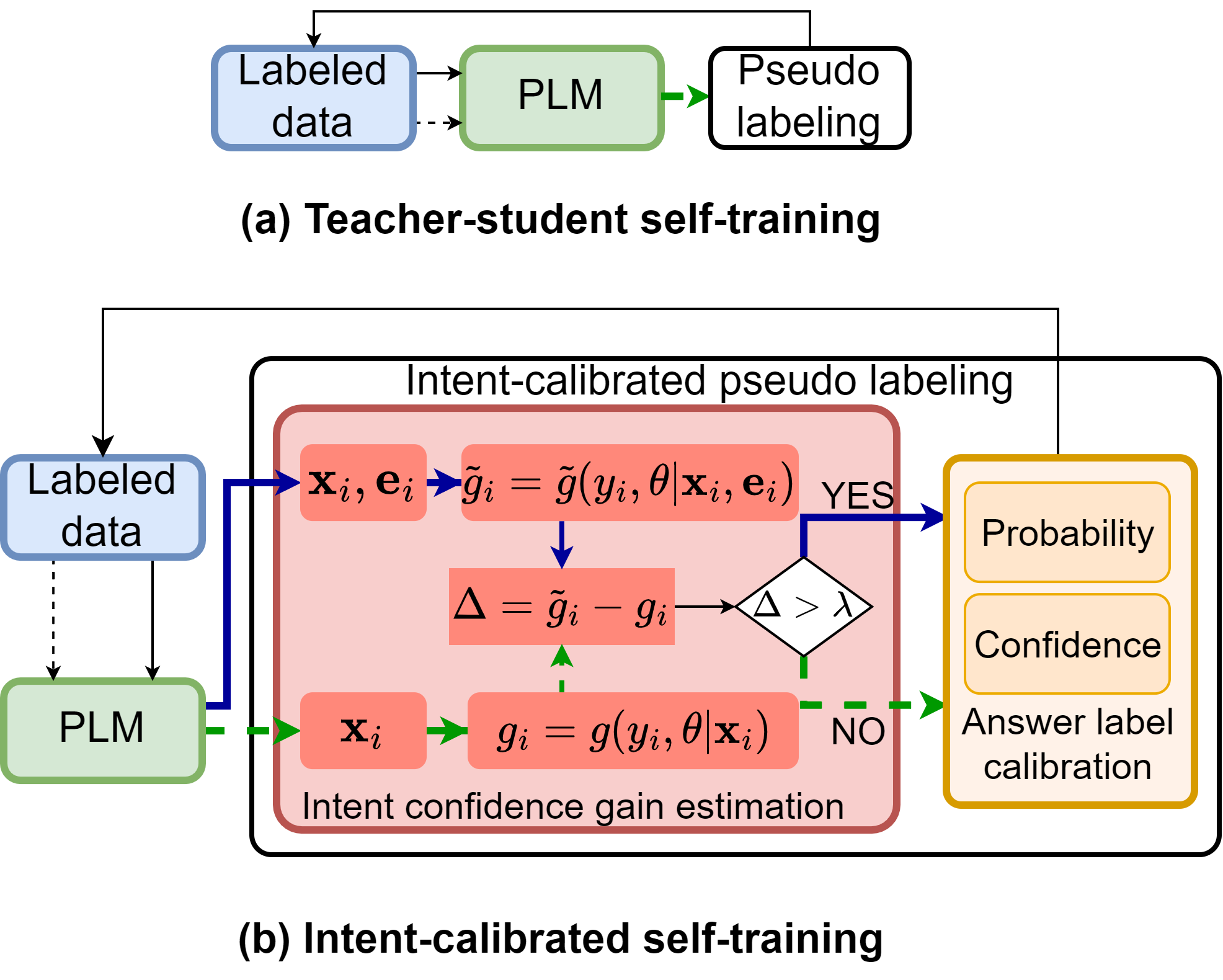}
    \caption{Comparison of self-training frameworks. (a) Teacher-student self-training framework. (b) Intent-calibrated self-training framework. The dashed thin line and solid thin line represent the workflow of teacher model and student model, respectively. The dashed thick line and solid thick line represent intent-aware and context-aware workflow.
    }
    \label{fig:method}
\end{figure}

\subsection{Teacher-student self-training framework}
The teacher-student self-training framework~\cite{li2021metats} is shown in Figure \ref{fig:method} (a). 
It first trains the teacher model with the labeled data $D^l$ to predict correct answer probabilities. 
Then at each iteration, the pseudo labeling module selects samples by using teacher's predictions to assign pseudo answer labels. 
Finally, the student model is trained with the labeled data and pseudo-labeled data. 
At the next iteration, the student model is used as a new teacher model. 

\noindent\textit{Pseudo labeling module}. 
The principle is to determine a subset of samples and assign the unlabeled samples with pseudo answer labels.
Following \citet{tur2005combining} and \citet{amini2022self}, we introduce thresholds $\lambda^+$ and $\lambda^-$ for the positive and the negative classes to select a subset of unlabeled data, with which the classifier is the most confident. 
For each unlabeled data, the selection criterion is defined as:
\begin{equation}\label{sample-selector}
d = (\exists! p_i > \lambda^{+})~ \mathord{?}~ 1 : 0,\\
\end{equation}
where $\exists!$ means exists one and only one. 
If there exists one and only one candidate answer, the probability $p_i$ (See Eq.~\ref{p-cls}) of which is larger than the positive threshold $\lambda^{+}$, then $d=1$ and we add the current sample to the subset for pseudo labeling.
Then, the pseudo answer label $y_i$ of each sample $\mathbf{x}_{i} \in X$ is assigned by:
\begin{align}
\label{pseudo-labeling}
y_i & =\Phi(\lambda^+,\lambda^-,p_i)= \left \{
\begin{array}{lr}
    1,     & p_i>\lambda^+\\
    0,     & p_i<\lambda^-
\end{array}
.
\right.
\end{align}
If the probability $p_i$ is sufficiently high ($p_i>\lambda^+$), then positive label ``1'' is assigned to $y_i$; 
if the probability $p_i$ is sufficiently low ($p_i<\lambda^-$), then negative label ``0'' is assigned to $y_i$;
otherwise $p_i \in [\lambda^-,\lambda^+]$, $y_i$ cannot be assigned a pseudo answer label, and this sample will not be used to train the student model. 

\begin{algorithm}[t!]

    \caption{\Acl{ICAST}}
    \label{icast_algorithm}
    \SetKwData{Left}{left}\SetKwData{This}{this}\SetKwData{Up}{up} \SetKwFunction{Union}{Union}\SetKwFunction{FindCompress}{FindCompress} \SetKwInOut{Input}{input}\SetKwInOut{Output}{output}
    	
    	\Input{Dataset w/ and wo/ labels $[D^l$,$D^u]$,
    	teacher model $f=[f^{\alpha},f^{\beta}]$, threshold $\lambda$ of intent confidence gain. 
     }
    	\Output{Student model $\tilde{f}$. }
    	\BlankLine
    	Train teacher model $f$ with $D^l$\;
    	\For{each $\mathbf{x}_i \in D^u$}{
    	   $\mathbf{e}_i \leftarrow f^{\alpha}(\mathbf{x}_i)$\tcp*[l]{Predict intents (Eq.\ref{igen})}
    	     $g_i\gets g(y_i, \beta \vert \mathbf{x}_i)$\tcp*[l]{Context-aware confidence (Eq.\ref{context-aware-confidence})}
    	   $\tilde{g}_i \gets \tilde{g}(y_i, \beta \vert \mathbf{x}_i, \mathbf{e}_i)$\tcp*[l]{Intent-aware confidence (Eq.\ref{intent-aware-confidence})}
              $\Delta=\tilde{g}_i-g_i$ \tcp*[l]{Intent confidence gain}
    	   \uIf{$\Delta>\lambda$}{
    	   $\tilde{\mathbf{x}}_i \leftarrow (\mathbf{x}_i,\mathbf{e}_i)$ \tcp*[l]{Update input with intents}
    	   }
    	   \Else{$\tilde{\mathbf{x}}_i \leftarrow \mathbf{x}_i$\;}
        }
    Select samples by Eq.~\ref{sample-selector} and Eq.~\ref{hard_data_selection1}\; 
    Assign a label for each sample by Eq.\ref{pseudo-labeling1} and collect a dataset $D^p=\{(\tilde{\mathbf{x}}_i, y_i)\}|_{i=1}^{|D^p|}$\;
    Re-train $\tilde{f}$ on $D^l\cup D^p$ with 5 epochs\;
    $f \leftarrow \tilde{f}$\tcp*[l]{Update teacher model with student model}
    Back to line 2 to iterate line 2-16 until the maximum epoch. 
\end{algorithm}

\section{Intent-calibrated Self-training}
\subsection{Overview}
We illustrate the proposed \Acf{ICAST}, as shown in Figure \ref{fig:method} (b). 
First, we train a teacher model on labeled data $D^l$ to predict pseudo intent labels for unlabeled data $D^u$ (See \S\ref{tmt}). 
Second, we conduct intent-calibrated pseudo labeling (See \S\ref{icpl}).
Specifically, we estimate intent confidence gain to select samples with high-quality intent labels, and we calibrate the answer labels by incorporating selected intent labels as an extra input for answer selection.
Third, we train the student model with labeled and pseudo-labeled data (See \S\ref{student-retraining}). 
We summarize the proposed intent-calibrated self-training in algorithm~\ref{icast_algorithm}.

\subsection{Teacher model training}
\label{tmt}
We first train a teacher model $f=[f^{\alpha},f^{\beta}]$ with the labeled dataset $D^{l}$.
The intent generation module $f^{\alpha}$ constructs its input as a sequence of tokens, i.e., $\mathbf{x}_i=[u_{1}; e_{1};\cdots;u_{\vert \mathbf{u}_{i} \vert};e_{\vert \mathbf{u}_{i} \vert};[SEP]]$.
It generates an intent label $e_j$ by computing the probability of a candidate intent label, where $j\in [1, \vert\mathbf{u}_i\vert]$:

\begin{align}
    p_i^{e}&=\sigma(W \mathbf{h}^{SEP}_i+b)\label{igen}, \\
    \mathbf{h}^{SEP}_i&=BERT(\mathbf{x}_{i})\label{igen-bert}. 
\end{align}

The answer selection module $f^{\beta}$ constructs its input as a sequence of tokens, i.e.,  $\mathbf{x}_i = [[CLS];u_1;e_1;\cdots;u_{\vert \mathbf{u}_{i} \vert};e_{\vert \mathbf{u}_{i} \vert};[SEP];a_i]]$ and computes the probability of a candidate answer as Eq.~\ref{p-cls} to decide if it is the correct answer.

\subsection{Intent-calibrated pseudo labeling}\label{icpl}
\subsubsection{Intent confidence gain estimation}

The intent-aware calibrator selects high-quality intent labels by estimating intent confidence gain. 
Intent confidence gain refers to the increase in confidence score after considering the predicted intents. 
A larger intent confidence gain indicates that the predicted intent can bring a greater increase in confidence score. 
We define the intent confidence gain as:
\begin{equation}\label{intent-confidence-gain}
\Delta=\tilde{g}(y_{i},\beta \vert \mathbf{x}_i,\mathbf{e}_i)-g(y_{i},\beta \vert \mathbf{x}_i), 
\end{equation}
where $\beta$ is the model parameters sampled by \ac{MC dropout}. 
The Eq.~\ref{intent-confidence-gain} is formulated as the difference of two terms. 
The first term $\tilde{g}(y_{i},\beta \vert \mathbf{x}_i,\mathbf{e}_i)$ is the confidence score of MC dropout with predicted intents, while the second term $g(y_{i},\beta \vert \mathbf{x}_i)$ is the confidence score of MC dropout alone. 
The difference of two terms refers to the increase in confidence score after considering the predicted intents.

The $g(y_{i},\beta \vert \mathbf{x})$ is the confidence score of \ac{MC dropout}~\cite{gal2017deep}, which measures the decrease in Shannon entropy of answer prediction after using MC dropout sampling, i.e., 
the difference between the entropy of posterior and the expectation of the entropy of posteriors with \ac{MC dropout}. 
Formally, it can be defined and approximated as:
\begin{equation}
    \label{context-aware-confidence}
    \begin{aligned}
    g&(y_i, \beta \vert \mathbf{x}_i) \\
    =&\mathbf{H}[y_i \vert \mathbf{x}_i]
-\mathbf{E}_{P(\beta)}[\mathbf{H}[y_i \vert \mathbf{x}_i;\beta]]\\
 \approx &\mathbf{H}[\mathbf{E}_{P(\beta)}[y_i \vert \mathbf{x}_i;\beta]]
-\mathbf{E}_{P(\beta)}[\mathbf{H}[y_i \vert \mathbf{x}_i;\beta]]\\
 \approx & -\frac{1}{T} \sum\limits_{t=1}^{T} {p_t} \cdot \log{\frac{1}{T} \sum\limits_{t=1}^{T} {p_t}} \\
        &+{\frac{1}{T} \sum\limits_{t=1}^{T} {p_t \cdot \log p_t}},
\end{aligned}
\end{equation}
where the $\mathbf{H}[\cdot \vert \cdot]$ is the Shannon entropy.
The confidence score of MC dropout is calculated by the difference of two terms: 
the first term is the Shannon entropy with MC dropout, 
and the second term is the mean value of Shannon entropy with multiple MC dropout samplings. 

Similarly, the confidence score of MC
dropout with predicted intents $\tilde{g}(y_{i},\beta \vert \mathbf{x}_i,\mathbf{e}_i)$ can be defined and approximated as:
\begin{equation}
    \begin{aligned}
    \tilde{g}&(y_i, \beta \vert \mathbf{x}_i,\mathbf{e}_i)\\
    =&\mathbf{H}[y_i \vert \mathbf{x}_i, \mathbf{e}_i]
    -\mathbf{E}_{P(\beta)}[\mathbf{H}[y_i \vert \mathbf{x}_i, \mathbf{e}_i;\beta]]\\
    \approx& \mathbf{H}[\mathbf{E}_{P(\beta)}[y_i \vert \mathbf{x}_i, \mathbf{e}_i;\beta]]
    -\mathbf{E}_{P(\beta)}[\mathbf{H}[y_i \vert \mathbf{x}_i, \mathbf{e}_i;\beta]]\\
    \approx& -\frac{1}{T} \sum\limits_{t=1}^{T} {p^{e}_{i}\tilde{p}_{t}} \cdot \log{\frac{1}{T} \sum\limits_{t=1}^{T} {p^{e}_{i} \tilde{p}_{t}}} \\
        &+{\frac{1}{T} \sum\limits_{t=1}^{T} {p^{e}_{i}\tilde{p}_t \cdot \log p^{e}_{i}\tilde{p}_t}},
    \label{intent-aware-confidence}
\end{aligned}
\end{equation}
the confidence score of MC dropout with predicted intents indicates the decrease of Shannon entropy of answer prediction after using MC dropout sampling with considering predicted intents. 
It includes the predicted intents as inputs to the model, which is different from Eq.~\ref{context-aware-confidence}. 

The first term of intent confidence gain is the confidence score of MC dropout after considering pseudo intents and the second term of intent confidence gain is the confidence score of MC dropout. 
The intent confidence gain is to measure how much confidence can pseudo intents can bring to the model with \ac{MC dropout}.  
The higher the intent confidence gain, the more improvement that predicted intents can bring to the confidence score. 
We set a threshold $\lambda$ to determine if the predicted intents can bring enough improvement to the confidence score. 
If the intent confidence gain is larger than $\lambda$, we conclude that the predicted intents can improve the confidence score sufficiently and add them to the model's inputs.
Specifically, if $\Delta > \lambda$, then we update the input with extra predicted intent labels $\mathbf{e}_i$, i.e.,  $\tilde{\mathbf{x}}_{i}=[\mathbf{x}_{i}, \mathbf{e}_{i}]$, 
which is expected to bring higher confidence score to the model, otherwise $\tilde{\mathbf{x}}_{i}=\mathbf{x}_{i}$.

\subsubsection{Answer label calibration}\label{answer-label-calibration}
To make use of more unlabeled samples, we introduce extra three thresholds $\tilde{\lambda}^{+}$, $\tilde{\lambda}^{-}$, and $\lambda^{h}$ to revise Eq.~\ref{sample-selector} as:
\begin{equation}\label{hard_data_selection1}
d = (\bar{p}_i > \tilde{\lambda}^{+}) \wedge (g_i > \lambda^{h}) ? 1 : 0,
\end{equation}
where $\lambda^{-}<\tilde{\lambda}^{-} \leq \tilde{\lambda}^{+} < \lambda^{+}$ and therefore we can consider extra samples with probabilities $p_{i} \in [\lambda^{-}, \lambda^{+}]$. 
The probability $\bar{p}_i$ is approximated by $T$ times MC dropout.
The threshold $\lambda^h$ is to select samples with high confidence $g_i$.
Formally, $\bar{p}_i$ and $g_i$ are defined as:
\begin{align}
\label{p-average}
\left \{
\begin{array}{ll}
    \bar{p}_{i} &= \frac{1}{T} \sum\limits_{t=1}^{T}{\tilde{p}_{t}p^{e}_{i}}, 
    g_i=\tilde{g}(y_i, \beta \vert \mathbf{x}_i,\mathbf{e}_i), 
    \Delta>\lambda\\
    \bar{p}_{i} &= \frac{1}{T} \sum\limits_{t=1}^{T}{p_{t}}, 
    g_i=g(y_i, \beta \vert \mathbf{x}_i),     
    \ \Delta \leq \lambda
\end{array}
.
\right.
\end{align}
To calibrate an answer label for each sample $\tilde{\mathbf{x}}_{i}$, we revise Eq.~\ref{pseudo-labeling} as:
\begin{align}
\label{pseudo-labeling1}
y_i=& \left \{
\begin{array}{lr}
    \Phi(\lambda^+,\lambda^-,\bar{p}_i),
    \bar{p}_{i}\in [0,\lambda^{-})\cup(\lambda^{+},1]\\
    \Phi(\tilde{\lambda}^+,\tilde{\lambda}^-,\bar{p}_i), 
    \bar{p}_{i} \in [\lambda^{-}, \lambda^{+}]
\end{array}
.
\right.
\end{align}
Afterward, we can get a pseudo labeled dataset $D^{p}=\{\tilde{\mathbf{x}}_i,y_i\}\vert_{i=1}^{\vert D^p \vert}$. 

Note that line 13 in Algorithm 1 shows the process of selecting pseudo answer labels for retraining the answer selection module of the student model: Eq.~\ref{sample-selector} is used to determine if we add a sample to the subset for pseudo labeling. Eq.~\ref{hard_data_selection1} is a revision of Eq.~\ref{sample-selector}, which aims to make use of more unlabeled samples by introducing extra three thresholds. 
The goal of selecting samples by criteria of Eq.~\ref{sample-selector} and Eq.~\ref{hard_data_selection1} (line 13 of Algorithm 1) is to prepare a set of candidates in primaries for high-quality pseudo labeling.

\subsection{Student model re-training}\label{student-retraining}

We re-train the student model $\tilde{f}=[\tilde{f}^{\alpha},\tilde{f}^{\beta}]$ with the extended dataset $D^{l}\cup D^{p}$. 
We minimize three types of binary cross entropy losses, i.e., intent generation loss $\mathcal{L}^{e}_{i}$, answer selection loss without intent labels $\mathcal{L}_{i}$ and answer selection loss with intent labels $\Tilde{\mathcal{L}}_{i}$, which are calculated as follows:  
\begin{align}
\begin{split}
&\mathcal{L}^{e}_{i}=-\mathbf{e}_{i}\log\tilde{f}^{\alpha}\mathbf{x}_{i} +(1-\mathbf{e}_{i})\log(1-\tilde{f}^{\alpha}(\mathbf{x}_{i})),\\
&\mathcal{L}_{i}=-y_{i}\log\tilde{f}^{\beta}\mathbf{x}_{i}+(1-y_{i})\log(1-\tilde{f}^{\beta}(\mathbf{x}_{i})),\\
&\Tilde{\mathcal{L}}_{i}=-y_{i}\log\tilde{f}^{\beta}(\mathbf{x}_{i},\mathbf{e}_{i})\\ &+(1-y_{i})\log(1-\tilde{f}^{\beta}(\mathbf{x}_{i},\mathbf{e}_{i})).
\end{split}
\end{align}
The intent generation loss $\mathcal{L}^{e}_{i}$ calculates the cross entropy loss between predicted intents and ground-truth intents. It can be used to optimize the intent generation module $\tilde{f}^{\alpha}$. 

The answer selection loss without intent labels $\mathcal{L}_{i}$ calculates the cross entropy loss between predicted answers and ground-truth answers. It can be used to optimize the answer selection module $\tilde{f}^{\beta}$ when the intent confidence gain is lower than the threshold.

The answer selection loss with intent labels $\mathcal{L}_{i}$ calculates the cross entropy loss between predicted answers and ground-truth answers. It can be used to optimize the answer selection module $\tilde{f}^{\beta}$ when the intent confidence gain is larger than the threshold.

\section{Experimental Setup}

\begin{table}[htp]
\caption{The statistics of experimental datasets, where labeled proportion denotes the proportion of labeled data in the training set. }\label{tab:data-stats}
\footnotesize
\centering
\setlength{\tabcolsep}{1.8pt}
\begin{tabular}{lrrrrr}
\toprule[2pt]
                          & \multicolumn{1}{c}{} & \multicolumn{2}{c}{Train}                                   & \multicolumn{1}{c}{Validation} & \multicolumn{1}{c}{Test} \\
                          & \multicolumn{1}{c}{} & \multicolumn{1}{r}{Labeled} & \multicolumn{1}{r}{Unlabeled} & \multicolumn{1}{c}{}           & \multicolumn{1}{c}{}     \\
\midrule[1.5pt]
\multirow{3}{*}{MSDIALOG} & 1\% & 1,410 & 140,420 & 5,000 & 21,280 \\
                          & 5\% & 7,050 & 134,780 & 5,000 & 21,280 \\
                          & 10\% & 14,100 & 127,730 & 5,000 & 21,280 \\ 
\midrule
\multirow{3}{*}{MANTIS}   & 1\% & 2,640 & 260,990 & 12,000 & 50,000 \\
                          & 5\% & 13,200 & 250,430 & 12,000 & 50,000 \\
                          & 10\% & 26,400 & 237,230 & 12,000 & 50,000 \\
\bottomrule[2pt]
\end{tabular}
\end{table}
\subsection{Datasets and evaluation metrics} \label{sec:dataset}
We test all methods on our extension of two benchmark datasets:
MSDIALOG~\cite{qu2018analyzing} and MANTIS~\cite{penha2019introducing}. 
MSDIALOG dataset contains multi-turn question answering across 4 topics collected from Microsoft community\footnote{\url{https://answers.microsoft.com}}.
It has 12 different types of intents. 
MANTIS dataset provides multi-turn dialogs with user intent labels across 14 domains crawled from stack exchange\footnote{\url{https://stackexchange.com/}\label{fn:stackex}}. 
It has 10 different types of intents. 
Note that we require a small number of data with intent labels in our experiments. 
There are other response selection datasets (e.g., UDC~\cite{lowe2015ubuntu}), however, they do not contain dialogues with intent labels. 
To this end, we select the MSDIALOG and MANTIS datasets which contain a small amount of data with intent labels, which can satisfy our experimental requirements.

Particularly, we extend both datasets with unlabeled data. 
For MSDIALOG, we treat data without intent labels as unlabeled data; for MANTIS, we crawl unlabeled data from Stack Exchange\footref{fn:stackex} from 2021 to 2022.
For fair comparison with baselines, we follow the previous work~\cite{zhangsam, yang2020iart, han2021fine}: we use the ground-truth label as positive sample and use BM25 algorithm~\cite{robertson2009probabilistic} to retrieve 9 relevant samples from different dialogues as negative samples. 
There could be a small number of negatives that are false-negatives, because there are cases which have same answers in different dialogues, so false-negatives may exist, but the number is very small. 
Besides, we use a different data partitioning strategy: 
First, we only extract conversations containing accurate answers, and the ground truth labels of all data are accurate answers.
This is because we focus on the answer selection task, while the prior works focus on the response selection task. 
Note that not all the responses can serve as answers to users’ questions. 
Second, we put the data with intent labels into the training set. This is because the number of the data with intent labels is small, and we want to fully utilize the intent labels. 
In order to compare different methods in low-resource settings, we design three different low-resource simulation experiments: including 1\%, 5\% and 10\% labeled data and a large amount of unlabeled data.
The statistics of the extended datasets is shown in Table~\ref{tab:data-stats}.  

We use 2 types of metrics to evaluate the models: classification metrics, i.e., Precision (P), Recall (R) and F1 score, and 
ranking metrics~\cite{yang2020iart, pan2021learning}, i.e.,  \ac{MAP} and Recall@k (R@k). 
\subsection{Baselines}
We compare the proposed \ac{ICAST} with recent state-of-the-art methods that have reported results on the MSDIALOG and MANTIS dataset, respectively.

\begin{itemize}[leftmargin=*,nosep]
\item \textbf{IART}~\cite{yang2020iart} proposes the intent-aware attention mechanism to weight the utterances in context. 
\item \textbf{SAM}~\cite{zhangsam} captures semantic and similarity features to enhance answer selection. 
\item \textbf{JM}~\cite{zhang2021once} concatenates the context and all candidate responses as input to select the most proper response. 
\item \textbf{BIG}~\cite{deng2021learning} uses the bilateral generation method to augment data and designs a contrastive loss function for training. 
\item \textbf{GRN}~\cite{liu2021graph} uses NUP and UOP pre-training tasks, and combines the graph network and sequence network to model the reasoning process of multi-turn response selection. 
\item \textbf{GRAY}~\cite{lin2020world} generates grayscale data by a fine-tuned generation model and proposes a multi-level ranking loss function for training. 
\item \textbf{BERT\_FP}~\cite{whang2020effective} learns the interactions between utterances in context to enhance answer selection. 
\item \textbf{BERT}~\cite{kenton2019bert} is a general classification framework, which predicts answer labels on the vector of $[CLS]$ token. 
\item \textbf{\Acf{TSST}}~\cite{li2021metats} is a semi-supervised method, a teacher model is first trained with small, labeled data to generate pseudo labels on a large unlabeled dataset and then train a student model with pseudo labels. 
\end{itemize}

\subsection{Implementation details}
All models are implemented based on Pytorch\footnote{\url{https://pytorch.org/}} and Huggingface\footnote{\url{https://huggingface.co/}}.
We conduct hyper-parameter tuning on the validation dataset and report results on the test dataset. 
We use BERT-base-uncased model~\cite{kenton2019bert} as an encoder in both $f^{\alpha}$ and $f^{\beta}$, where the parameters are shared. 
We use the AdamW~\cite{loshchilov2017decoupled} as optimizer.
The batch size is 16, initial learning rate is 5e-5, and weight decay is 0.01. 
The maximum number of context turn is set to 4. 
The maximum length of context and answer are set to 400 and 100. 
The dropout ratios is 0.1. 
\Ac{ICAST} generates pseudo labels every 5 epochs. 
\ac{MC dropout} conducts sampling by $T=5$ times. 
For thresholds of pseudo labeling, we set $\lambda^{+}=0.8$, $\lambda^{-}=0.1$, ${\tilde\lambda}^{+}=0.5$, $\tilde{\lambda}^{-}=0.5$ and $\lambda^{h}=0.2$.
For the threshold of intent confidence gain, we set $\lambda=0.0$ for MANTIS dataset with 5\% and 10\% labeled data, otherwise, $\lambda=0.02$. 

For each parameter, we fix other hyper-parameters and select a specific value for the best performance on validation datasets. 
$\lambda^-$, $\lambda^+$ and $\lambda^h$ are selected in (0, 1), the grid is 0.1. 
$\tilde{\lambda}^-$ and ${\tilde{\lambda}^+}$ are selected in (0.1, 0.8), the grid is 0.1. 
$\tilde{\lambda}$ is selected in $[0, 0.05)$, the grid is 0.01. 
The number of \ac{ICAST}'s parameters is 109,493,005. 
We train \ac{ICAST} on 2 2080Ti GPUs with random seed 42, and the time cost is 48 hours.

\begin{table*}[ht]
\caption{Overall performance of answer selection.
Bold and underlined fonts indicate leading and compared results in each setting.
1\%, 5\% and 10\% are the proportion of labeled data in training dataset. 
The symbol $\dagger$ indicates the baselines reproduced by the released source codes and $\ddagger$ indictates the baselines we implemented based on the papers. Note that we cannot fairly compare with the reported results in the IART paper, because we use a different data partitioning for a different task (See Section~\ref{sec:dataset}).}
\label{tab:overall_result}
\resizebox{\linewidth}{!}{ 
\begin{tabular}{llcccccccccccccc}
\toprule[2pt]
 & & \multicolumn{7}{c}{\text{MSDIALOG}} & \multicolumn{7}{c}{\text{MANTIS}} \\  
 \cmidrule(r){3-9}  \cmidrule(r){10-16}
Setting   & Model & P & R & F1 & R@1 & R@2 & R@5 & MAP & P & R & F1 & R@1 & R@2 & R@5 & MAP \\ 
\midrule[1.5pt]
 \multicolumn{1}{l}{\multirow{5}{*}{\tabincell{l}{1\% labeled}}} & IART$^\dagger$ & 22.18 & 46.75 & 30.08 & 25.65 & 46.28 & 77.58 & 47.74 & 48.29 & 52.22 & 50.18 & 50.40 & 68.34 & 86.22 & 66.12 \\
                     \multicolumn{1}{l}{}                   & SAM$^\dagger$ & 44.17 & 44.36 & 44.26 & 46.89 & 59.06 & 77.02 & 60.72 & 57.75 & 58.62 & 58.18 & 65.10 & 76.32 & 88.54 & 75.60  \\
                     \multicolumn{1}{l}{}                   & JM$^\ddagger$ & 44.80 & 44.59 & 44.70 & 44.54 & 60.76 & 84.30 & 61.26 & 62.95 & 62.62 & 62.78 & 62.64 & 77.32 & 92.30 & 75.25  \\
                     \multicolumn{1}{l}{}                   & BIG$^\ddagger$ & 44.07 & 44.78 & 44.42 & 50.93 & 66.30 & 87.50 & 66.15 & 57.91 & 57.42 & 57.66 & 70.22 & 83.04 & 95.12 & 80.78  \\
                     \multicolumn{1}{l}{}                   & GRAY$^\ddagger$ & 41.68 & 42.15 & 41.91 & 51.26 & 66.40 & 85.62 & 66.10 & 61.30 & 60.72 & 61.01 & 64.67 & 77.34 & 88.32 & 75.57  \\
                     \multicolumn{1}{l}{}                   & GRN$^\ddagger$ & 43.41 & 43.37 & 43.39 & 43.28 & 61.60 & 86.46 & 61.19 & 61.75 & 61.10 & 61.42 & 61.06 & 76.64 & 93.66 & 74.56  \\
                     \multicolumn{1}{l}{}                   & BERT\_FP$^\dagger$ & 44.32 & 42.95 & 43.62 & 56.76 & \textbf{72.08} & \textbf{91.25} & \textbf{70.90} & 66.26 & 62.86 & 64.51 & 75.62 & \textbf{86.14} & \textbf{95.22} & \textbf{84.11} \\
                     \multicolumn{1}{l}{}                   & BERT$^\ddagger$ & 48.56 & 45.34 & 46.90 & 54.79 & 68.32 & 85.80 & 68.04 & 67.28 & 65.62 & 66.44 & 74.82 & 83.00 & 92.16 & 82.41 \\ 
                     \multicolumn{1}{l}{}                   & ICAST (Teacher) & \textbf{49.82} & \textbf{46.33} & \textbf{48.01} & \textbf{56.86} & 67.81 & 85.38 & 69.03 & \textbf{68.48} & \textbf{66.12} & \textbf{67.28} & \textbf{77.28} & 86.12 & 94.98 & 82.98 \\ 
\midrule
 \multicolumn{1}{l}{\multirow{2}{*}{\tabincell{l}{1\% labeled \\ +all unlabeled}}} & TSST$^\ddagger$ & 53.72 & 52.58 & 53.14 & 61.04 & 73.91 & 89.70 & 73.04 & 73.73 & 72.60 & 73.16 & 82.94 & \textbf{91.08} & \textbf{97.88} & \textbf{89.18} \\
                     \multicolumn{1}{l}{}                   & ICAST & \textbf{57.05} & \textbf{54.32} & \textbf{55.65} & \textbf{62.21} & \textbf{76.31} & \textbf{91.07} & \textbf{73.77} & \textbf{74.89} & \textbf{72.72} & \textbf{73.79} & \textbf{83.68} & 90.68 & 96.42 & 88.31 \\
\midrule[1.5pt]
                     \multicolumn{1}{l}{\multirow{5}{*}{\tabincell{l}{5\% labeled}}} & IART$^\dagger$ & 23.52 & 49.38 & 31.86 & 28.80 & 48.02 & 79.93 & 49.97 & 50.24 & 53.60 & 51.86 & 51.56 & 70.66 & 89.52 & 67.75 \\
                     \multicolumn{1}{l}{}                   & SAM$^\dagger$ & 49.52 & 51.45 & 50.47 & 54.27 & 67.66 & 83.03 & 67.32 & 59.16 & 57.82 & 58.48 & 66.52 & 76.88 & 89.28 & 76.51 \\
                     \multicolumn{1}{l}{}                   & JM$^\ddagger$ & 50.98 & 49.81 & 50.39 & 50.37 & 67.62 & 89.47 & 66.56 & 67.16 & 66.82 & 66.99 & 66.92 & 80.83 & 94.94 & 78.47  \\
                     \multicolumn{1}{l}{}                   & BIG$^\ddagger$ & 50.82 & 50.93 & 50.88 & 58.12 & 73.07 & 89.80 & 71.53 & 61.34 & 60.88 & 61.11 & 74.22 & 87.58 & 96.94 & 84.02  \\
                     \multicolumn{1}{l}{}                   & GRAY$^\ddagger$ & 48.99 & 48.26 & 48.62 & 55.16 & 69.54 & 86.23 & 68.75 & 62.53 & 66.50 & 64.45 & 70.24 & 80.92 & 90.62 & 79.48  \\
                     \multicolumn{1}{l}{}                   & GRN$^\ddagger$ & 49.28 & 50.04 & 49.66 & 49.76 & 66.77 & 89.52 & 66.04 & 64.27 & 63.00 & 63.62 & 63.78 & 78.60 & 93.38 & 76.27  \\
                     \multicolumn{1}{l}{}                   & BERT\_FP$^\dagger$ & 49.74 & 50.93 & 50.33 & \textbf{62.96} & \textbf{77.16} & \textbf{92.76} & \textbf{75.41} & 70.04 & 68.32 & 69.17 & 80.22 & \textbf{89.36} & \textbf{97.30} & \textbf{87.37} \\
                     \multicolumn{1}{l}{}                   & BERT$^\ddagger$ & 52.01 & 49.67 & 50.81 & 61.23 & 72.60 & 85.19 & 72.13 & 71.17 & 67.56 & 69.32 & 77.70 & 86.82 & 95.48 & 85.20 \\
                     \multicolumn{1}{l}{}                   & ICAST (Teacher) & \textbf{54.22} & \textbf{51.83} & \textbf{53.00} & 62.59 & 74.38 & 90.36 & 74.16 & \textbf{73.13} & \textbf{69.20} & \textbf{71.11} & \textbf{80.82} & 88.12 & 95.86 & 84.27 \\ 
\midrule
 \multicolumn{1}{l}{\multirow{2}{*}{\tabincell{l}{5\% labeled \\ +all unlabeled}}} & TSST$^\ddagger$ & 58.34 & 58.78 & 58.56 & 64.89 & 74.62 & 86.41 & 74.61 & 74.33 & 72.92 & 73.61 & 81.62 & 89.32 & 96.10 & 87.83 \\
                     \multicolumn{1}{l}{}                   & ICAST & \textbf{61.54} & \textbf{59.72} & \textbf{60.62} & \textbf{69.54} & \textbf{80.35} & \textbf{93.09} & \textbf{77.29} & \textbf{74.60} & \textbf{74.62} & \textbf{74.61} & \textbf{84.38} & \textbf{90.76} & \textbf{97.06} & \textbf{89.78} \\ 
\midrule[1.5pt]
                     \multicolumn{1}{l}{\multirow{5}{*}{\tabincell{l}{10\% labeled}}} & IART$^\dagger$ & 34.38 & 47.22 & 39.79 & 39.05 & 58.31 & 84.77 & 58.00 & 50.77 & 53.04 & 51.88 & 51.80 & 71.20 & 89.28 & 68.04 \\
                     \multicolumn{1}{l}{}                   & SAM$^\dagger$ & 55.63 & 54.27 & 54.94 & 59.53 & 70.62 & 85.05 & 71.00 & 61.39 & 60.00 & 60.69 & 66.88 & 77.92 & 90.84 & 77.08 \\
                     \multicolumn{1}{l}{}                   & JM$^\ddagger$ & 57.64 & 57.56 & 57.60 & 57.61 & 73.12 & 90.97 & 71.70 & 68.06 & 68.98 & 68.52 & 68.22 & 80.46 & 94.04 & 79.02  \\
                     \multicolumn{1}{l}{}                   & BIG$^\ddagger$ & 56.15 & 55.96 & 56.06 & 62.96 & 76.78 & 90.08 & 74.92 & 62.74 & 62.34 & 62.54 & 76.60 & 87.62 & 96.28 & 85.08  \\
                     \multicolumn{1}{l}{}                   & GRAY$^\ddagger$ & 54.46 & 53.05 & 53.75 & 62.45 & 76.08 & 90.60 & 74.51 & 65.20 & 65.26 & 65.23 & 74.80 & 85.74 & 94.66 & 83.51  \\
                     \multicolumn{1}{l}{}                   & GRN$^\ddagger$ & 54.06 & 53.43 & 53.74 & 53.52 & 70.67 & 90.08 & 68.96 & 66.01 & 64.92 & 65.46 & 66.10 & 80.60 & 93.34 & 77.83  \\
                     \multicolumn{1}{l}{}                   & BERT\_FP$^\dagger$ & 57.95 & 56.81 & 57.38 & \textbf{67.48} & \textbf{80.16} & \textbf{94.07} & \textbf{78.56} & 71.04 & 68.20 & 69.59 & 80.72 & 89.38 & 96.82 & 87.59 \\
                     \multicolumn{1}{l}{}                   & BERT$^\ddagger$ & 61.94 & \textbf{60.19} & 61.05 & 64.38 & 73.77 & 85.99 & 74.12 & 70.33 & 69.56 & 69.94 & \textbf{82.12} & \textbf{91.00} & \textbf{97.70 }& \textbf{88.72} \\
                     \multicolumn{1}{l}{}                   & ICAST (Teacher) & \textbf{62.41} & \underline{59.77} & \textbf{61.06} & 66.54 & 76.55 & 89.09 & 76.43 & \textbf{71.89} & \textbf{70.24} & \textbf{71.05} & 81.92 & 90.02 & 97.12 & 88.29 \\ 
\midrule
 \multicolumn{1}{l}{\multirow{2}{*}{\tabincell{l}{10\% labeled \\ +all unlabeled}}}  & TSST$^\ddagger$ & 63.28 & 63.34 & 63.31 & 70.91 & \textbf{81.95} & \textbf{93.18} & \textbf{80.37} & 76.17 & 73.34 & 74.73 & 83.70 & 91.18 & \textbf{97.50} & \textbf{89.43} \\
                     \multicolumn{1}{l}{}                   & ICAST & \textbf{65.98} & \textbf{64.89} & \textbf{65.43} & \textbf{72.27} & \textbf{81.95} & 91.63 & 79.63 & \textbf{77.43}& \textbf{73.36} & \textbf{75.35} & \textbf{84.60} & \textbf{91.52} & 97.36  & 88.59 \\ 
\bottomrule[2pt]

\end{tabular}
}
\end{table*}
\section{Results}
\subsection{Overall performance}

We compare the overall performance of \ac{ICAST} against the baseline methods. 
We also report the results of ICAST(Teacher). 
ICAST(Teacher) uses intent labels but BERT and BERT\_FP do not, which is not a fair comparison.
We conduct these experiments to see whether our method can outperform baselines without using unlabeled data. 
The results of overall performance are shown in Table~\ref{tab:overall_result}.

First, in terms of all classification metrics, ICAST and ICAST (Teacher)  outperform the baselines in each setting, excluding only one setting: the R score of ICAST (Teacher) is 0.42\% lower than BERT trained on 10\% labeled MSDIALOG dataset. 
Specifically, on MSDIALOG dataset, ICAST with 1\%, 5\%, and 10\% labeled data improves their corresponding strongest baselines by 2.51\%, 2.06\%, and 2.12\% of F1 scores.
On MANTIS dataset, ICAST with 1\%, 5\%, and 10\% labeled data improves their corresponding strongest baselines by 0.63\%, 1.00\%, and 0.62\% of F1 scores.
This demonstrates the effectiveness of \ac{ICAST} on the performance of classifying correct answers.
We believe there are two reasons:
\begin{enumerate*}[nosep, label={(\roman*)}]
\item the predicted intent labels can provide more information that are useful for selecting correct answers; and
\item the self-training paradigm can calibrate answer labels for continuous improvement. For example, with self-training on 10\% labeled MSDIALOG dataset and all unlabeled data, the R score of ICAST is 4.70\%/1.55\% higher than BERT/TSST, respectively. 
\end{enumerate*}

Second, in terms of ranking metrics, we have the following observations: 
\begin{enumerate*}[nosep, label={(\roman*)}]
\item ICAST outperforms all baselines in terms of R@1 score in each setting.
Specifically, on MSDIALOG dataset, ICAST with 1\%, 5\%, and 10\% labeled data achieve 1.17\%, 4.65\%, and 1.36\% higher of R@1 scores than their corresponding strongest baselines, respectively. 
On MANTIS dataset, ICAST with 1\%, 5\%, and 10\% labeled data achieve 0.74\%, 2.76\%, and 0.90\% higher of R@1 scores than their corresponding strongest baselines, respectively. 
It indicates that ICAST can rank an accurate answer on top. 
\item For R@2, R@5 and MAP, ICAST achieves highest scores in most of the settings, excluding: on MSDIALOG dataset, with 10\% labeled data and all unlabeled data, R@5 and MAP scores decrease 1.55\% and 0.74\%. On MANTIS dataset, with 1\% labeled data and all unlabeled data, R@2, R@5 and MAP scores decrease 0.40\%, 1.46\% and 0.87\%; with 10\% labeled data and all unlabeled data, R@5 and MAP scores decrease 0.14\% and 0.84\%.
Our method does not possess a significant advantage in terms of R@2, R@5, and MAP, as the primary objective of answer selection is to identify the answer rather than generating a ranking list. Hence, the fundamental performance measurements are precision, recall, and the F1-score~\cite{wang2021comqa}. Accordingly, we evaluate the models using standard evaluation metrics for a fair comparison. Additionally, we present supplementary ranking metrics (i.e., R@2, R@5, MAP) to assess whether improvements in selection metrics result in a noteworthy decline in ranking metrics. The results demonstrate that our method exhibits no considerable decrease in ranking metrics.
\end{enumerate*}

Third, using self-training with unlabeled data has large impact on all settings in terms of both classification and ranking metrics.
Specifically, on MSDIALOG dataset with 1\%, 5\%, and 10\% labeled data, F1 scores increase 7.64\%, 7.62\%, 4.37\%; MAP scores increase 2.87\%, 1.88\%, and 1.81\%.
On MANTIS dataset with 1\%, 5\%, and 10\% labeled data, F1 scores increase 6.51\%, 3.50\%, and 4.30\%; MAP scores increase 5.07\%, 2.41\%, and 0.71\%.
This reveals that ICAST benefits from making good use of unlabeled data with self-training.
Besides, the influence of classification performance is larger than the ranking performance in each setting. 

Last but not least, we do not require too much data with intent labels. This also motivates us to conduct experiments on only a small amount of data with labels (1\%, 5\%, 10\%).
For example, with 1\% labeled data, our method outperforms the baselines with only 141 and 264 intent labels on MSDIALOG and MANTIS datasets respectively. 
Thus, it is possible to apply our method in practice, even without a large amount of intent labels.

\subsection{Ablation study}

\begin{table*}[ht]
\caption{Ablation study. Impact of different modules in our proposed framework. $\myUparrow$ and $\myuparrow$ indicates an increase of the performance compared with \acs{ICAST} and \acs{ICAST}-\acs{ICGE}, respectively. 
}\label{ablation}
\resizebox{\linewidth}{!}{ 
\begin{tabular}{llcccccccccccccc}
\toprule[2pt]
 & & \multicolumn{7}{c}{MSDIALOG} & \multicolumn{7}{c}{MANTIS} \\  \cmidrule(r){3-9}  \cmidrule(r){10-16}
Setting & Model & P & R & F1 & R@1 & R@2 & R@5 & MAP & P & R & F1 & R@1 & R@2 & R@5 & MAP \\ 
\midrule[1.5pt]
\multicolumn{1}{l}{\multirow{4}{*}{\tabincell{l}{1\% labeled \\ +all unlabeled}}} & ICAST & {57.05} & {54.32} & {55.65} & {62.21} & {76.31} & {91.07} & {73.77} & {74.89} & {72.72} & {73.79} & {83.68} & {90.68} & {96.42} & {88.31} \\ \cmidrule(r){2-16}
                     \multicolumn{1}{l}{}                   & -\acs{ICGE} & 54.81 & 54.04 & 54.42 & 60.24 & 72.85 & 86.46 & 71.83 & 74.44 & 71.78 & 73.08 & 81.68 & 89.02 & 94.86 $\myUparrow$ & 87.55 \\
                     \multicolumn{1}{l}{}                   & -\acs{ICGE}-\acs{ALC} & 54.13 & 53.52 & 53.82 & 60.80 $\myuparrow$ & 73.77 $\myuparrow$ & 89.38 $\myuparrow$ & 72.86 $\myuparrow$ & 74.00 & 71.96 $\myuparrow$ & 72.96 & 82.66 $\myuparrow$ & 90.82 $\myUparrow$ $\myuparrow$ & 97.72 $\myUparrow$ $\myuparrow$ & 88.94 $\myUparrow$ $\myuparrow$ \\
                     \multicolumn{1}{l}{}                   & -\acs{IG} & 53.72 & 52.58 & 53.14 & 61.04 & 73.91 & 89.70 & 73.04 & 73.73 & 72.60 & 73.16 & 82.94 & {91.08} $\myUparrow$ & {97.88} $\myUparrow$ & {89.18} $\myUparrow$ \\ 
\midrule[1.5pt]
                     \multicolumn{1}{l}{\multirow{4}{*}{\tabincell{l}{5\% labeled \\ +all unlabeled}}} & ICAST & {61.54} & {59.72} & {60.62} & {69.54} & {80.35} & {93.09} & {77.29} & {74.60} & {74.62} & {74.61} & {84.38} & 90.76 & 97.06 & {89.78} \\ \cmidrule(r){2-16}
                     \multicolumn{1}{l}{}                   & -\acs{ICGE} & 60.66 & 58.55 & 59.58 & 64.94 & 76.31 & 90.32 & 75.61 & 74.35 & 74.12 & 74.23 & 83.18 & {91.14} $\myUparrow$  & {97.50} $\myUparrow$ & 89.29 \\
                     \multicolumn{1}{l}{}                   & -\acs{ICGE}-\acs{ALC} & 59.94 & 58.92 $\myuparrow$ & 59.43 & 66.30 $\myuparrow$ & 77.82 $\myuparrow$ & 91.63 $\myuparrow$ & 76.86 $\myuparrow$ & 74.19 & 73.56 & 73.87 & 82.80 & 90.72 & 96.66 & 88.81 \\
                     \multicolumn{1}{l}{}                   & -\acs{IG} & 58.34 & 58.78 & 58.56 & 64.89 & 74.62 & 86.41 & 74.61 & 74.33 & 72.92 & 73.61 & 81.62 & 89.32 & 96.10 & 87.83 \\ 
\midrule[1.5pt]
                     \multicolumn{1}{l}{\multirow{4}{*}{\tabincell{l}{10\% labeled \\ +all unlabeled}}} & ICAST & {65.98} & {64.89} & {65.43} & {72.27} & {81.95} & 91.63 & 79.63 & {77.46} & 73.36 & {75.35} & {84.60} & {91.52} & {97.36}  & 88.59 \\ \cmidrule(r){2-16}
                     \multicolumn{1}{l}{}                   & -\acs{ICGE} & 65.71 & 62.78 & 64.21 & 71.42 & 81.53 & {93.70} $\myUparrow$ & {80.51} $\myUparrow$ & 76.19 & 73.82 $\myUparrow$ & 74.98 & 83.82 & 91.26 & 96.98 & 89.46 $\myUparrow$\\
                     \multicolumn{1}{l}{}                   & -\acs{ICGE}-\acs{ALC} & 64.27 & 64.09 & 64.18 $\myuparrow$ & 70.63 & {81.95} & 92.95 & 80.15 & 75.60 & {74.26} & 74.92 & 84.06 & 91.38 & 97.32 & {89.69} \\
                     \multicolumn{1}{l}{}                   & -\acs{IG} & 63.28 & 63.34 & 63.31 & 70.91 & {81.95} & 93.18 $\myUparrow$ & 80.37 $\myUparrow$ & 76.17 & 73.34 & 74.73 & 83.70 & 91.18 & {97.50} $\myUparrow$ & 89.43 $\myUparrow$ \\ 
\bottomrule[2pt]

\end{tabular}
}
\end{table*} 
To better understand the contribution of each functional components of ICAST i.e., \acf{ICGE}, \acf{ALC}, and \acf{IG}, we conduct the following ablation studies: 
After removing the IG (denoted as ``-IG''), ICGE and ALC do not work, so ICAST degenerates to the TSST model. 
The ICGE estimates intent confidence gain according to the predicted intents which are the outputs of the IG module. The ALC uses the intent confidence scores to select unlabeled samples, the computation of intent confidence scores also needs the predicted intents. 
After removing the ICGE (denoted as ``-ICGE''), the model does not select the predicted intents according to the intent confidence gain and adds all of predicted intent labels into the inputs.
After removing the ICGE and ALC (denoted as ``-ICGE-ALC''),  the model does not select unlabeled data with intent confidence score and degenerates into the TSST with adding all predicted intents into inputs. 
Table~\ref{ablation} reports the results of the ablation studies.

First, \acf{ICGE}, \acf{ALC}, and \acf{IG} have positive influence on overall performance of classification in all settings on both MSDIALOG and MANTIS datasets with 1\%, 5\%, and 10\% labeled data, respectively.
Removing \ac{IG} from \ac{ICAST}, F1 scores decrease 2.51\%/2.06\%/2.12\% on MSDIALOG dataset and 0.63\%/1.00\%/0.62\% on MANTIS dataset.
This proves our hypothesis that the generated intents can provide more useful information for selecting correct answers.
Removing \ac{ICGE} from \ac{ICAST}, F1 scores decrease 1.23\%/1.04\%/1.22\% on MSDIALOG dataset and 0.71\%/0.38\%/0.37\% on MANTIS dataset.
This reveals that intent confidence gain can select high-quality intent labels that are helpful to select correct answers.
Removing \ac{ALC} from \ac{ICAST} without \ac{ICGE}, F1 scores decrease 0.60\%/0.15\%/0.03\%.
This shows that \ac{ALC} can bring extra improvement even though \ac{ICGE} is absent. 
Meanwhile, it works better together with the other two components.

Second, in terms of ranking performance, R@1 decreases when removing \ac{ICGE}, \ac{ALC}, and \ac{IG} from \ac{ICAST} in all settings on MSDIALOG and MANTIS datasets.
Removing \ac{ICGE}/\ac{ALC}/\ac{IG} with 1\%, 5\%, and 10\% labeled data, R@1 drops 1.97\%/1.41\%/1.17\%, 4.60\%/3.24\%/4.65\%, and 0.85\%/1.64\%/1.36\% on MSDIALOG dataset; 
R@1 drops 2.00\%/1.02\%/0.74\%, 1.20/1.58/2.76\%, and 0.78\%/0.54\%/0.90\% on MANTIS dataset.
This shows that the three functional components are helpful to rank correct answers on top.

\subsection{Analysis}

\begin{figure}[htbp]
    \centering
    \includegraphics[width=\columnwidth]{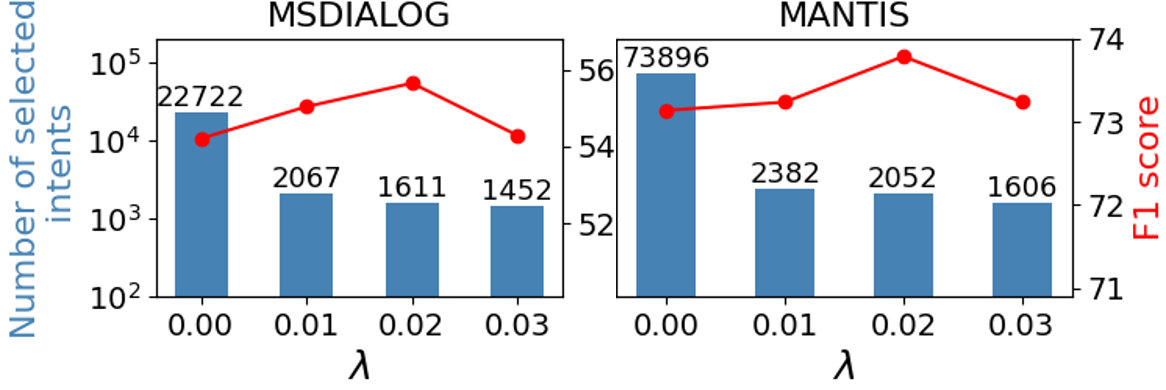}
    \caption{
    F1 scores (w.r.t. the line) and average number of selected intents for answer label calibration (w.r.t. the bar) with different values of $\lambda=[0.00, 0.01, 0.02, 0.03]$ on MSDIALOG (left) and MANTIS (right) with 1\% labeled data.
    }
    \label{fig:analysis}
\end{figure}

Figure~\ref{fig:analysis} shows the impact of threshold of intent confidence gain $\lambda$ on classification performance of \ac{ICAST}.
We can see that as $\lambda$ increases, average number
of selected intents decreases, meanwhile, F1 scores increase first, achieve top at $\lambda=0.02$ and then descend. 
Because with a larger $\lambda$, more intents are selected to calibrate answer labels, which leads to an increase of F1 scores.
Then adding more generated intents might also introduce noise for answer selection, which is the possible reason for decrease of F1.
Thus, $\lambda$ can balance between more predicted intents and less noisy intents.

\subsection{Case study}
\begin{table}[t]
\setlength{\tabcolsep}{2pt}
\centering
\caption{A case study of selecting answers by \ac{ICAST} and \ac{TSST} model. 
Each model chooses the candidate answer with the highest probability among all candidate answers as the correct answer. 
If \ac{ICG} is larger than $\lambda$, then \ac{ICAST} combines the intents and context to select answer. 
Here, $\lambda=0.00$. 
Note that the first candidate answer (A1) is the correct answer.
}\label{tab:case_study}
\resizebox{\linewidth}{!}{ 
\begin{tabular}{@{}llccc@{}}
\toprule[2pt]
\multicolumn{4}{l}{Context Utterances}  & \begin{tabular}[c]{@{}c@{}} Intents \end{tabular} \\ \midrule
\multicolumn{4}{l}{\begin{tabular}[c]{@{}l@{}}User: How does a photon picture make the pattern?
\end{tabular}}                  & \multicolumn{1}{c}{OQ}                      \\\specialrule{0em}{3pt}{3pt}
\multicolumn{4}{l}{\begin{tabular}[c]{@{}l@{}}Agent: Photons in mainstream physics, are quantum\\ mechanical entities which in great numbers build up\\ the classical electromagnetic radiation...
\end{tabular}}                 & \multicolumn{1}{c}{PA}                      \\\specialrule{0em}{3pt}{3pt}
\multicolumn{4}{l}{\begin{tabular}[c]{@{}l@{}}User: Do you know why the photon which is hitting\\ forward is causing an electron to move up-down?
\end{tabular}}                  & \multicolumn{1}{c}{IR}                   \\ \specialrule{0em}{1.5pt}{1.5pt} \midrule \specialrule{0em}{0.5pt}{0.5pt} \midrule
\multicolumn{2}{l}{Candidate Answers}  &  Model  & \acs{ICG} & Probablity  \\ \midrule
\multicolumn{2}{l}{\multirow{3}{*}{\begin{tabular}[l]{@{}l@{}}A1: The theories of quantum mechanics\\ for electron photon interactions can\\ be found in https://www.website.com. \end{tabular}}}  & \ac{TSST}& / & 0.00    \\
\multicolumn{2}{l}{} &  &  & \\
\multicolumn{2}{l}{} & \ac{ICAST} & 0.14 & 0.99\\ 
\midrule\specialrule{0em}{3pt}{3pt}
\multicolumn{2}{l}{\multirow{3}{*}{\begin{tabular}[l]{@{}l@{}} A2: The energy of a photon is equal to\\ the level spacing of a two-level system.\\ It is a result of energy conservation...\end{tabular}}} & TSST & / & 0.96 \\
\multicolumn{2}{l}{} &  &  & \\
\multicolumn{2}{l}{} & \ac{ICAST} & -0.13 & 0.71\\

\bottomrule[2pt]
\end{tabular}
}
\end{table}

Table~\ref{tab:case_study} shows a case study of how \ac{ICAST} and \ac{TSST} select different answers for the same given context.

In general, a model chooses the candidate answer with the highest probability among all candidate answers as the correct answer.
In this case, the strongest baseline \ac{TSST} incorrectly chooses the second candidate answer (A2) with the probability of 0.96, instead of the first candidate answer (A1) which has a probability of 0.00. 
It shows that selecting answers based solely on their probabilities can result in significant bias.
\Ac{ICAST} calibrates the probabilities based on \ac{ICG}, and it correctly chooses A1 with the probability of 0.99, while skip A2 with the probability of 0.71.
\Ac{ICAST} computes ICGs by combining context and its predicted intents, and each candidate answer. 
The ICG of correct answer is larger than $\lambda$,  which indicates that \ac{ICAST} can capture the intent information from the correct answer, so \ac{ICAST} increases the probability of correct answer from 0.00 to 0.99. 
Meanwhile, the ICG of incorrect answer is less than $\lambda$, which indicates \ac{ICAST} cannot capture the intent information from incorrect answer, so \ac{ICAST} decreases the probability of incorrect answer from 0.96 to 0.71. 
Furthermore, we explain the intuition.
In context utterances, the user asks the \acf{OQ}, and then the agent gives a \acf{PA} which can explain the original question, but the user still raises the \acf{IR} to ask the agent for more detailed information. 
Next, the user anticipates an answer that includes a link or document providing more detailed information, instead of a continuous explanation in text.
Intuitively, the predicted intents can aid in monitoring changes in the user's expectations throughout the utterances.
\section{Conclusion and Future Work}

In this paper, we propose the \acf{ICAST} based on the teacher-student self-training and intent-calibrated answer selection: 
we train a teacher model on labeled data to predict intent labels on unlabeled data; 
select high-quality intents by intent confidence gain to enrich inputs and predict pseudo answer labels; 
and retrain a student model on both the labeled and pseudo-labeled data.
We conduct massive experiments on two benchmark datasets and 
the results show that \ac{ICAST} outperforms baselines even with small but same proportion (i.e., 1\%, 5\% and 10\%) of labeled data, respectively. 
Note that we understand more proportion of labeled data may lead to an increase of performance, e.g., BERT-FP with 10\% labeled data beat ICAST with 1\% labeled data for across all metrics for MSDIALOG.
However, we focus on verifying if the proposed \ac{ICAST} outperforms other methods given a very few amounts of labeled data.
In some cases, \ac{ICAST} can outperform baselines with fewer labeled data. 
In the future work, we will explore more predictable dialogue context (e.g., user profiles) than intents.

\section{Reproducibility}
To facilitate reproducibility of the results reported in this paper, the code and data used are available at \url{https://github.com/dengwentao99/ICAST}. 

\section*{Limitation}
Our proposed \ac{ICAST} also has the following limitations. 
First, \ac{ICAST} only considers user intents to enhance answer selection. 
It is limited because we only capture the user's expectations from the predicted intent labels, without considering other user-centered factors, such as user profiles and user feedback. 
Second, like retrieval-based methods which have been shown to have a good effect in professional question-answering fields, \ac{ICAST} also has limitations when it comes to diversity.
For example, \ac{ICAST} cannot retrieve multiple correct answers with different expressions given the same context.
Third, since our model needs to predict the intent labels, to complete this task , the model needs a few additional parameters.

\section*{Ethical Considerations}
We realize that there are risks in developing the dialogue system, so it is necessary to pay attention to the ethical issues of the dialogue system. 
It is crucial for a dialogue system to give correct answers to the users while avoiding ethical problems such as privacy preservation problems. 
Due to the fact that we have used the public datasets to train our model, these datasets are carefully processed by publishers to ensure that there are no ethical problems. 
Specifically, the dataset publishers performed user ID anonymization on all datasets, and only the tokens ``user'' and ``agent'' are used to represent the roles in the conversation.
The utterances do not contain any user privacy information (e.g., names, phone numbers, addresses) to prevent privacy disclosure.

\section*{Acknowledgement}
We would like to thank the editors and reviewers for their helpful comments.
This research was supported by the National Key R\&D Program of China with grant (No.2022YFC3303004, No.2020YFB1406704), the Natural Science Foundation of China (62102234, 62272274, 62202271, 61902219, 61972234, 62072279), the Key Scientifc and Technological Innovation Program of Shandong Province (2019JZZY010129), the Fundamental Research Funds of Shandong University, and VOXReality (European Union grant, No. 101070521). 
All content represents the opinion of the authors, which is not necessarily shared or endorsed by their respective employers and/or sponsors.

\bibliography{contents/references}
\bibliographystyle{acl_natbib}

\end{document}